\documentclass[10pt,twocolumn,letterpaper]{article}

\usepackage{cvpr}              
\definecolor{cvprblue}{rgb}{0.21,0.49,0.74}
\usepackage[pagebackref,breaklinks,colorlinks,allcolors=cvprblue]{hyperref}

\def\paperID{*****} 
\def\confName{CVPR}
\def\confYear{2026}
\graphicspath{{figures/}}
\newcommand{\dataset}{VRR-QA}
\newcommand{\method}{Adaptive Dense Evidence Refinement}

\title{\vspace{-0.4in}\method{} for Video Relational Reasoning for \dataset{} Challenge}
\author{
Yuyang Sun\textsuperscript{1},
Yongliang Wu\textsuperscript{1},
Xingyu Zhu\textsuperscript{2},
Yuxia Chen\textsuperscript{3}, \\
Zhenxiang Jiang\textsuperscript{4},
Yangguang Ji\textsuperscript{4},
Wenbo Zhu\textsuperscript{4},
Yanxi Shi\textsuperscript{4},
Jay Wu\textsuperscript{4},
Shuo Wang\textsuperscript{5},
Xu Yang\textsuperscript{1} \\
\textsuperscript{1}Southeast University
\textsuperscript{2}National University of Singapore
\textsuperscript{3}Independent Researcher \\
\textsuperscript{4}Opus AI Research
\textsuperscript{5}University of Science and Technology of China‌
}

\begin{document}
\maketitle
\begin{abstract}
\dataset{} evaluates whether video-language systems can infer spatial,
temporal, viewpoint, depth, and visibility relations that are not always
resolved by a single frame. We present an inference-only system built around
adaptive test-time computation. The system first answers each question with a
direct video-language model pass, then uses multiple lightweight views to find
unstable questions. Only these difficult questions are routed to a high-budget
dense evidence module that constructs timestamped frame observations,
relation-specific probes, candidate verification, and conservative
temporal aggregation. This design separates two problems that are often
confused in video question answering: finding plausible alternative answers and
deciding when a current answer should actually be changed. On the test split,
the final system obtains 90.07 average accuracy and 87.81 macro average
accuracy. The report focuses on the final test system and the implementation
settings required to reproduce the adaptive dense verifier.
\end{abstract}
    
\section{Introduction}

Video question answering benchmarks increasingly test whether multimodal
models can reason over time rather than only recognize visible objects or
actions \cite{fu2025video}. \dataset{} focuses on visual
relational reasoning beyond explicit cues: a question may require inferring
which object is closer, where an entity is located relative to another entity,
whether an object is visible from a viewpoint, or how a relation changes across
the clip \cite{swetha2025vrrqa}. These questions are difficult because the decisive
evidence is often distributed across frames and may depend on the correct
reference frame.

Our system uses no supervised training or fine-tuning. It is an inference-time
pipeline that treats video relation reasoning as an evidence allocation
problem. A single direct video-language model pass is efficient and often
strong, but it compresses all visual evidence into one answer. This is risky
for relation questions where the model must distinguish screen coordinates from
object-centric coordinates, separate object motion from camera motion, or
resolve an occlusion from only a few frames.

We therefore allocate more computation only when the question appears unstable.
Multiple lightweight views produce candidate answers and disagreement signals.
Stable questions keep the direct answer. Difficult questions are sent to a dense
evidence refinement module that makes the visual reasoning explicit through
timestamped observations, relation probes, and pairwise verification between
the current answer and a candidate alternative.

The main contribution is a conservative refinement framework for video
relational reasoning. The method does not treat every model disagreement as a
correction. Instead, it uses disagreement to identify hard cases, then requires
visible temporal evidence before changing the answer. This keeps the method
reproducible and explains why additional model calls are spent on a small set
of high-risk questions rather than uniformly across the full test set.

\section{Method}

\subsection{Overview}

\begin{figure*}[t]
\centering
\includegraphics[width=0.98\linewidth]{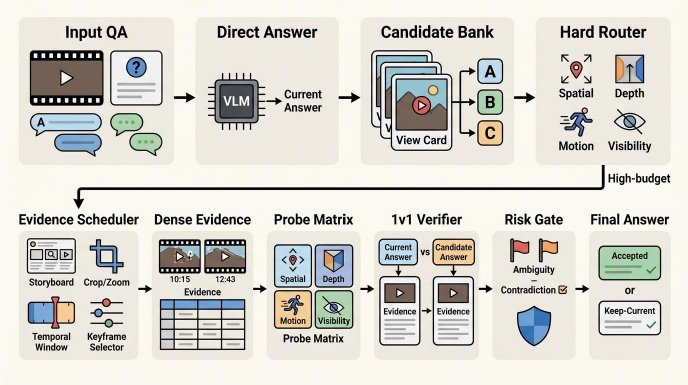}
\caption{\method{} pipeline. A direct answer is first produced for all
questions. Multi-view evidence builds a candidate bank and identifies unstable
relation questions. A high-budget branch then schedules dense frame evidence,
relation probes, pairwise verification, and risk-gated temporal aggregation
before any answer is changed.}
\label{fig:method}
\end{figure*}

Figure~\ref{fig:method} shows the full pipeline. The method is an adaptive
test-time reasoning system rather than a single prompt. It separates answer
production into four nested layers: a direct answer layer, a candidate bank
layer, an evidence construction layer, and a risk-controlled decision layer.
The direct layer is applied to all questions. The later layers are activated
only when the answer is unstable or when the question asks for a relation that
is likely to require explicit temporal evidence.

\subsection{Direct Video Inference}

The base predictor receives the full video, question, and answer options, and
returns a single answer choice. This direct path is the default inference path
for all questions. It is simple, reproducible, and avoids unnecessary answer
changes when the visual evidence is already clear.

Direct inference is not discarded by later stages. It defines the current
answer that every candidate must challenge. This replacement framing is
important: a candidate answer is accepted only if it is better supported by
visual evidence than the current answer, not merely because it is plausible.

\subsection{Candidate Bank Construction}

Before verification, the system builds a compact candidate bank. Each question
is evaluated under several lightweight views of the same video-question pair.
The views differ in frame sampling, storyboard construction, crop or zoom
emphasis, temporal alignment, answer-option ordering, and relation-focused
prompting. Each view returns an answer and a short evidence statement.

The bank stores the current answer, non-current answer candidates, supporting
views, relation keywords, and evidence snippets. It is intentionally
recall-oriented: a candidate may enter the bank even if it is not yet safe to
apply. Candidate generation and answer replacement are treated as separate
problems.

\subsection{Disagreement and Hard-Case Routing}

The router decides whether a question should remain on the direct path or enter
the expensive evidence path. It combines answer instability, candidate support,
relation type, and answer-option structure. Questions receive higher priority
when independent views disagree, when a non-current answer appears repeatedly,
or when the question type is likely to require frame-level evidence:

\begin{itemize}
    \item lateral, vertical, and front/back spatial relations;
    \item relative depth and proximity;
    \item motion direction and trajectory changes;
    \item viewpoint, visibility, facing direction, and occlusion.
\end{itemize}

The routing signal is not used as a final answer predictor. It only decides the
amount and type of computation assigned to the question. Stable questions keep
the direct answer, mild disagreements receive pairwise verification, and hard
relation questions receive dense evidence construction.

\subsection{Adaptive Evidence Scheduler}

The high-budget branch uses a fixed sequence of modules, but the number of
visual views depends on the routed difficulty. The scheduler starts with a
global evidence scout, then expands into dense frame chunks only when the scout
finds relevant entities and candidate timestamps. For relation questions, it
also requests targeted crops, local zooms, and temporal windows around the
suspected decisive moments.

\begin{table}[t]
\centering
\caption{Adaptive evidence schedule for difficult questions. The high-budget
path is deliberately more expensive than direct inference, but it is activated
only for selected unstable cases.}
\label{tab:evidence-schedule}
\scriptsize
\setlength{\tabcolsep}{3pt}
\begin{tabular}{p{0.29\linewidth}p{0.30\linewidth}p{0.30\linewidth}}
\toprule
Stage & Evidence produced & Purpose \\
\midrule
Scout & entities, rough timestamps & localize relevant moments \\
Storyboard views & global frame sequence & preserve temporal context \\
Crop/zoom views & local entity evidence & resolve small objects and occlusion \\
Temporal windows & before/after relations & check motion and state changes \\
Probe matrix & relation-specific facts & test spatial, depth, motion, visibility \\
Pairwise verifier & current-vs-candidate decision & test replacement safety \\
Risk gate & ambiguity and contradiction flags & block unsafe corrections \\
\bottomrule
\end{tabular}
\end{table}

\subsection{Dense Frame Evidence Refinement}

For a selected difficult question, the video is converted into a sequence of
timestamped evidence chunks. Instead of asking the model to immediately choose
an answer again, the refinement stage asks for structured observations: which
entities are visible, which reference object matters, what relation is visible,
which timestamps support each hypothesis, and which ambiguity flags are present.

This produces an explicit evidence table over time. The final decision can
therefore be traced to visual observations rather than to a single global model
judgment.

\subsection{Relation Probe Matrix}

The verifier uses a probe matrix specialized to the predicted relation type.
Spatial probes check left/right, front/behind, above/below, and reference-frame
consistency. Depth probes check foreground/background, occlusion, and proximity.
Motion probes compare start and end positions and separate object motion from
camera motion. Visibility probes inspect line of sight, facing direction, and
blocking objects. Counterfactual probes ask what visual evidence would make the
current answer impossible and whether that evidence is present in the clip.

These probes make the reasoning narrower and easier to audit. They also reduce
the chance that a visually similar but unsupported candidate wins simply because
it matches the broad scene.

\subsection{Pairwise and Counterfactual Verification}

For each candidate correction, the verifier compares exactly two hypotheses:
the current answer and the candidate answer. The prompt asks which hypothesis is
better supported by the video evidence and requires evidence for both sides.
The verifier is not asked to freely choose any option at this stage.

The pairwise format turns answer refinement into a replacement test. A
candidate must identify a visual failure in the current answer and provide
timestamped support for itself. If the evidence is weak or ambiguous, the
system keeps the current answer.

\subsection{Risk-Gated Temporal Aggregation}

Frame-level observations are aggregated before any answer is changed. A
candidate is accepted only when multiple timestamped observations support it,
the current answer is contradicted by the evidence, the support is stable over
time or concentrated in a decisive temporal segment, and no high-risk ambiguity
flag is present. Otherwise, the direct answer is retained.

The risk gate blocks corrections when the reference object is ambiguous, the
relation depends on a single unclear frame, camera motion may explain the
apparent trajectory, the relevant entity is partially occluded, or the candidate
is supported only by a generic scene match. This conservative rule is critical
because raw model disagreement is noisy. The method uses disagreement to focus
computation, but it uses dense visual evidence to decide whether a correction is
safe.

\subsection{Inference-Time Implementation}

The system uses no supervised training or fine-tuning. The input is a video
clip, a question, and a fixed set of answer options. The direct path produces
one current answer for every question. The disagreement detector then evaluates
selected lightweight views, including uniformly sampled storyboards,
query-guided frames, crop or zoom views, and temporal-alignment prompts.

For hard cases, the dense refinement path builds a timestamped contact sheet and
selects relevant frames or chunks. Each chunk is converted into structured
fields: visible entities, reference object, observed relation, current-answer
support, candidate-answer support, and uncertainty. The final output remains a
standard answer file, while the intermediate records form an audit trail for the
small set of expensive cases.

\section{Experiments}

\subsection{Implementation}

The system is zero-shot and uses Gemini 3.1 Pro Preview ~\cite{gemini} for both direct
answering and evidence verification. Direct video calls use temperature 0.0,
top-$p$ 0.95, a 2,000-token thinking budget, and 256 output tokens. Evidence
calls use 12 uniformly sampled storyboard frames arranged in 4 columns with
320-pixel frame width; visual-review calls use temperature 0.2, top-$p$ 0.95, a
4,000-token thinking budget, and 1,536 output tokens. All answer-producing
calls are constrained to the provided options.

\subsection{Evaluation Protocol}

\dataset{} evaluates multiple-choice video relational reasoning. The system
outputs exactly one answer option for each question, and the evaluator reports
average accuracy and macro average accuracy. We report the final test
configuration only. Additional diagnostic runs are not used as main evidence
unless they are executed with the same end-to-end inference protocol as the
reported system.

This reporting choice keeps the experiment section aligned with the actual
test setting: the direct path is inexpensive and broad, while the dense evidence
path is a high-budget route reserved for selected hard cases.

\subsection{Test Results}

Table~\ref{tab:main-results} reports the test performance of the final system.
Scores are percentages. The final system uses direct video inference for stable
questions and applies the dense evidence verifier only to routed hard cases.

\begin{table}[t]
\centering
\caption{Test performance of the final system.}
\label{tab:main-results}
\small
\setlength{\tabcolsep}{4pt}
\begin{tabular}{ccc}
\toprule
Stage & AvgAcc & MacroAvgAcc \\
\midrule
\textbf{Final system} &  \textbf{90.07} & \textbf{87.81} \\
\bottomrule
\end{tabular}
\end{table}

\subsection{Test Ablation}

Table~\ref{tab:ablations} compares direct Gemini prompting with evidence-routed
systems on the test split. The direct baseline asks the multimodal model to
answer from the video context in one pass. The middle configuration verifies
candidate answers before replacement, but does not use the full dense evidence
schedule. The final system adds dense evidence refinement for routed hard cases.
This comparison isolates the value of explicit evidence construction around the
same Gemini backbone, while keeping the final output format unchanged.

\begin{table}[t]
\centering
\caption{Test ablation of evidence routing with the same Gemini backbone.}
\label{tab:ablations}
\scriptsize
\setlength{\tabcolsep}{3pt}
\begin{tabular}{ccc}
\toprule
Configuration & Inference policy & AvgAcc \\
\midrule
Direct Gemini prompting & One video pass & 70.47 \\
Simple verification & Candidate check & 82.18 \\
\textbf{Final system} & \textbf{Dense evidence refinement} & \textbf{90.07} \\
\bottomrule
\end{tabular}
\end{table}

\section{Conclusion}

We presented \method{}, an inference-only pipeline for \dataset{}. The method
uses direct video-language inference for stable questions and reserves
high-budget dense evidence analysis for difficult disagreement cases. The key
idea is to separate difficult case discovery from answer correction: multiple
views identify unstable questions, while dense timestamped evidence and
pairwise verification decide whether a candidate should replace the current
answer. This risk-controlled design is well suited to video relational
reasoning, where many errors depend on spatial reference frames, temporal
segments, depth, motion, and visibility.

{
    \small
    \bibliographystyle{ieeenat_fullname}
    \bibliography{main}

@article{gemini,
  title={Gemini: a family of highly capable multimodal models},
  author={Team, Gemini and Anil, Rohan and Borgeaud, Sebastian and Alayrac, Jean-Baptiste and Yu, Jiahui and Soricut, Radu and Schalkwyk, Johan and Dai, Andrew M and Hauth, Anja and Millican, Katie and others},
  journal={arXiv preprint arXiv:2312.11805},
  year={2023}
}

@inproceedings{fu2025video,
  title={Video-mme: The first-ever comprehensive evaluation benchmark of multi-modal llms in video analysis},
  author={Fu, Chaoyou and Dai, Yuhan and Luo, Yongdong and Li, Lei and Ren, Shuhuai and Zhang, Renrui and Wang, Zihan and Zhou, Chenyu and Shen, Yunhang and Zhang, Mengdan and others},
  booktitle={Proceedings of the IEEE/CVF conference on computer vision and pattern recognition},
  pages={24108--24118},
  year={2025}
}

@article{swetha2025vrrqa,
  title={VRR-QA: Visual Relational Reasoning in Videos Beyond Explicit Cues},
  author={Swetha, Sirnam and Gupta, Rohit and Kulkarni, Parth Parag and Shatwell, David G and Santiago, Jeffrey A Chan and Siddiqui, Nyle and Fioresi, Joseph and Shah, Mubarak},
  journal={arXiv preprint arXiv:2506.21742},
  year={2026}
}
}

\end{document}